\documentclass[conference]{IEEEtran}
\IEEEoverridecommandlockouts
\usepackage{cite}
\usepackage{amsmath,amssymb,amsfonts}
\usepackage{algorithmic}
\usepackage{graphicx}
\usepackage{textcomp}
\usepackage{xcolor}
\usepackage{times}
\usepackage{epsfig}
\usepackage{microtype}
\usepackage{graphicx}
\usepackage{subfigure}
\usepackage{booktabs} 
\usepackage{amsmath,amssymb,amsfonts}
\usepackage{algorithm}
\usepackage{textcomp}
\usepackage{xcolor}
\usepackage{soul}
\usepackage{makecell}
\usepackage{ctable}
\usepackage{booktabs} 
\usepackage{multirow}
\usepackage{thm-restate}
\usepackage{sistyle}
\SIthousandsep{,}
\usepackage{calligra}
\usepackage{enumitem}
\usepackage{subfigure}
\usepackage{footmisc} 
\usepackage{cite}
\usepackage{url}
\usepackage{wrapfig}

\def\BibTeX{{\rm B\kern-.05em{\sc i\kern-.025em b}\kern-.08em
    T\kern-.1667em\lower.7ex\hbox{E}\kern-.125emX}}

\begin{document}
\newcommand{\ours}[0]{\text{MGNet}}

\title{Multiplex Graph Networks for Multimodal Brain Network Analysis}

\makeatletter
\newcommand{\linebreakand}{%
  \end{@IEEEauthorhalign}
  \hfill\mbox{}\par
  \mbox{}\hfill\begin{@IEEEauthorhalign}
}
\makeatother

\author{
  \IEEEauthorblockN{Zhaoming Kong}
  \IEEEauthorblockA{\textit{Computer Science and Engineering} \\
    \textit{Lehigh University}\\
    zhk219@lehigh.edu}
  \and
  \IEEEauthorblockN{Lichao Sun}
  \IEEEauthorblockA{\textit{Computer Science and Engineering} \\
    \textit{Lehigh University}\\
    james.lichao.sun@gmail.com}
  \and
  \IEEEauthorblockN{Hao Peng}
  \IEEEauthorblockA{\textit{Cyber Science and Technology} \\
    \textit{Beihang University}\\
    penghao@act.buaa.edu.cn}
  \linebreakand 
  \IEEEauthorblockN{Liang Zhan}
  \IEEEauthorblockA{\textit{Electrical and Computer Engineering} \\
    \textit{University of Pittsburgh}\\
    liang.zhan@pitt.edu}
  \and
  \IEEEauthorblockN{Yong Chen}
  \IEEEauthorblockA{\textit{Perelman School of Medicine} \\
    \textit{University of Pennsylvania}\\
    ychen123@pennmedicine.upenn.edu}
  \and
  \IEEEauthorblockN{Lifang He}
  \IEEEauthorblockA{\textit{Computer Science and Engineering} \\
    \textit{Lehigh University}\\
    lih319@lehigh.edu}
}



\maketitle

\begin{abstract}
In this paper, we propose MGNet, a simple and effective multiplex graph convolutional network (GCN) model for multimodal brain network analysis. The proposed method integrates tensor representation into the multiplex GCN model to extract the latent structures of a set of multimodal brain networks, which allows an intuitive 'grasping' of the common space for multimodal data. Multimodal representations are then generated with multiplex GCNs to capture specific graph structures. We conduct classification task on two challenging real-world datasets (HIV and Bipolar disorder), and the proposed MGNet demonstrates state-of-the-art performance compared to competitive benchmark methods. Apart from objective evaluations, this study may bear special significance upon network theory to the understanding of human connectome in different modalities. The code is available at \url{https://github.com/ZhaomingKong/MGNets}.
\end{abstract}


\section{Introduction}
In recent years, brain network analysis has attracted considerable interests in the fields of neuroscience and machine learning. 
It plays a vital role in understanding biologically fundamental mechanisms of brain function, such as how the brain manages cognition, what signals the connections convey and how these signals affect brain regions \cite{fornito2013graph}. It has been found to be helpful in the early diagnosis of several neurological disorders, such as HIV infection and Alzheimer's disease \cite{fornito2016fundamentals, myszczynska2020applications}. However, databases for biomedical engineering are significantly smaller than for other applications, which make this study very challenging. It is a common belief that the fusion of complementary information from multimodal data would be highly beneficial to improve the effectiveness of brain network analysis. For example, functional magnetic resonance imaging (fMRI) and diffusion tensor imaging (DTI) are the most widely used modalities which can reflect morphological differences of brain regions and yield complementary information of brain networks. \\
\indent In the analysis of multimodal brain networks, various machine learning methods have been investigated for representation learning and disease prediction from shallow to deep models, such as multiple graph kernels \cite{jie2016sub}, tensor factorizations \cite{cao2017t,liu2018multi,he2018boosted}, and convolution neural networks \cite{plis2018reading,wang2017structural,kawahara2017brainnetcnn}.
Although significant progress has been made in this endeavor, there still lacks a basic model that can predict cutting performance. In particular, the brain network has sophisticated and non-linear structure, which may not be well captured by shallow models. Meanwhile, deep learning methods may suffer from excessive parameters, which are both difficult for training and vulnerable to overfitting. Besides, most existing methods do not make good use or even fail to accommodate the intrinsic graph structure of brain networks.\\
\indent Recently, graph convolutional networks (GCNs) have emerged as a promising approach to learn powerful representations for graph data and have been successfully used for brain network analysis \cite{li2019graph, arslan2018graph, ktena2018metric, cosmo2020latent, LiBrainGNN}. However, GCN explicitly requires a known graph structure (i.e., adjacency matrix), which is typically not available in brain networks, especially in multimodal cases. To address this challenge, Zhang et al. \cite{zhang2018multi} proposed a MVGCN method for single modality multi-view brain network classification, where prior knowledge of geometric coordinate is required to define a common graph structure. In addition, the small-world model was introduced to learn a random graph during training \cite{zhang2019new}. To the best of our knowledge, there is no general method for multimodal brain network analysis with GCNs.\\
\indent In this paper, we propose MGNet, a generic framework for multimodal brain network analysis by integrating tensor representation into multiplex GCN models. Different from existing methods \cite{zhang2018multi, LiBrainGNN} that rely on prior information such as regions of interest (ROI) for graph structure modeling, MGNet does not require any prior knowledge. It first stacks all multimodal brain networks across subjects and modalities into a 4D tensor, and then employs multilinear tensor projection to leverage the structural information of the 4D tensor to directly generate the graph structure. 
The main contributions of the paper are summarized as follows:

\begin{itemize} 
    \item We present a novel multimodal brain network classification framework (MGNet) by combining tensor and GCN techniques, where a multilinear tensor projection is introduced to take advantages of the homogeneity of multimodal data and the nature of graphs, and a graph structure is learned by projection pursuit. Meanwhile, an associated tensor form of GCN is proposed for combining multiple GCN models. 
    \item Our model can be regarded as a generalization of GCN to multi-graph data with tensor representation. It alleviates the problem of pre-determining population graph structure, which is often ambiguous due to the intrinsic complexity of intra-graph and cross-graph.
    \item We examine the effectiveness of the proposed MGNet model on three challenging real datasets (i.e., HIV, Bipolar and PPMI) for the disease prediction tasks, and the results demonstrate that MGNet model can produce state-of-the-art performance compared to strong baseline methods.
\end{itemize}

\section{Related Work} \label{sec:related}

In this section, we briefly introduce existing research related to this work.

\textbf{Tensor-based Embedding.}
Several tensor-based methods have been proposed to leverage the convenience of tensor representation to characterize multimodal graph data. Specifically, \cite{cao2017t} proposed an iterative tensor-based method for simultaneous graph embedding and clustering. \cite{ma2017multi} proposed a tensor-based embedding method for classification task with some multi-view side information. In \cite{liu2018multi}, partial symmetric tensor decomposition is applied to learn the multi-view multi-graph embeddings. Recently, \cite{yin2016low} introduced the $t$-product tensor factorization to handle multimodal and high-dimensional data. \\
\indent \textbf{Graph Convolutional Networks.}
In recent years, several CNN architectures for learning over graphs have been proposed. Specifically, \cite{kawahara2017brainnetcnn} introduced a novel convolution filter to leverage the topological locality of structural brain networks. \cite{ren20173} extended CNN to 3D functional brain network classification and \cite{wang2017structural} to structural brain network mining with graph reordering. More recently, \cite{arslan2018graph} explored GCN for the task of ROI identification and \cite{ktena2018metric} extended GCN for brain network pairwise-metric learning. \cite{zhang2018multi} proposed a MVGCN method for multi-view brain network classification, where prior knowledge of geometric coordinate is required to define a common feature representation space. In addition, \cite{zhang2019new} proposed to learn a random graph during training based on the small-world model, and \cite{LiBrainGNN} designed a ROI-aware brain graph model to take advantage of the topological and functional information of fMRI. \\
\indent We notice that most of the existing methods are customized for a specific purpose, and there still lacks a general GCN-based method for the multimodal and cross-modal brain network analysis and classification task.

\section{Methodology}\label{sec:method}
\subsection{Problem Formulation}
A brain network usually leverages a graph structure to describe interconnections between brain regions, which can be represented by a weighted graph $G = \{V, E, \mathbf{X}\}$, where $V = \{v_i\}_{i = 1}^N$ is the node set indicating brain regions, $E = \{e_{ij}\}_{i,j = 1}^N$ is the edge set between nodes, and $\mathbf{X} =\{x_{ij}\}_{i,j = 1}^N$ is the weighted matrix where $x_{ij}$ is the corresponding edge weight. Assume that each subject consists of $M$-modal brain networks $\{G_1, \cdots, G_M\}$, where each one is extracted from a specific imaging modality (or measure) such as fMRI and DTI. These networks share the same set of nodes, i.e., using an identical definition of brain regions, but may differ in network topology and edge weight. Given a multimodal brain network data set of $S$ subjects with $M$ different modalities $\mathcal{D} = \{ (\{G_{1s}, \cdots, G_{ms}, \cdots, G_{Ms} \}, y_s) \}_{s = 1}^S$, where $G_{ms}$ is the brain network data of the $m$-th modality in the $s$-th subject and $y_s$ the corresponding class label, the goal of multimodal brain network data analysis is to probe the interrelationships between different modalities and obtain from low-level or raw relational data higher-level descriptions of brain-behavior states to facilitate disease diagnosis or treatment monitoring.
\subsection{Architecture of {\ours}}
\begin{figure*}[htbp]
  \centering
    \includegraphics[width=0.9\linewidth]{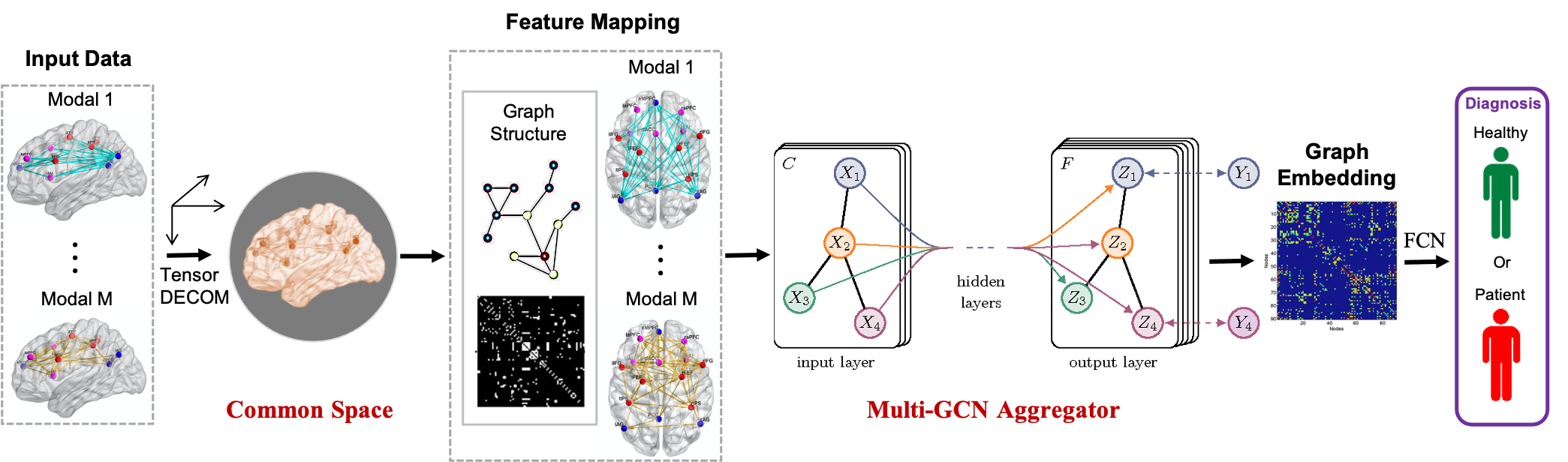}\\
  \caption{The architecture of {\ours}. It first conducts higher-order SVD decomposition of tensor data to study the group-wise maps of multimodal brain network fusion. Then multiple GCN aggregators jointly learn the representations of brain networks by passing the information in an inter-graph or intra-graph way. After that, the extracted features are fed to a fully connected layer (FCN) for prediction in an end-to-end manner.}
  \label{Fig_framework}
\end{figure*}
\vspace{-3pt}
Fig.~\ref{Fig_framework} provides an overview of the proposed tensor-driven multiplex graph convolutional network ({\ours}) model, which consists of two major components. Briefly, the first part performs cross-modality bridging with tensor decomposition, which is used to efficiently extract latent structures across modalities and individuals. The second part is the design phase of {\ours} model, which adopts GCN to capture intrisic data graph structures and encode relationship of different modalities. {\ours} can be used in both spectral and spatial domains, as there is an equivalence of the graph convolution process regardless of the specific domain \cite{balcilar2020bridging}. In this study, we consider a general multi-layer GCN model with the following propagation rule \cite{kipf2016semi}:
\begin{equation}
\label{eq:GCN}
    \mathbf{H}^{(l+1)} = \sigma(\widetilde{\mathbf{D}}^{-\frac{1}{2}} \widetilde{\mathbf{A}} \widetilde{\mathbf{D}}^{-\frac{1}{2}} \mathbf{H}^{(l)} \mathbf{W}^{(l)})
\end{equation}
where $\widetilde{\mathbf{A}} = \mathbf{A} + \mathbf{I}_N$ is the adjacency matrix of the undirected graph $G$ with self-connections $\mathbf{I}_N$, $\widetilde{d}_{ii} = \sum_j  \widetilde{a}_{ij}$ and $\mathbf{W}^{(l)}$ is a layer-specific trainable weight matrix. $\sigma(\cdot)$ denotes an activation function, such as the $ReLU(\cdot) = \max(0, \cdot)$, and $\mathbf{H}^{(l)} \in \mathbb{R}^{N \times D}$ is the feature matrix of the $l$-th layer.\\
\indent In the following, we discuss each component in detail and discuss how the form of this propagation rule can be effectively extended to multiplex models with tensor representation, thereby enabling the formation of a multimodal learning environment.
\subsubsection{Cross-Modality Bridging}
The brain network classification problem can be reformulated as a graph classification task, and many effective graph-centric models \cite{nguyen2019universal, hamilton2017representation} can be utilized by treating each modality independently and specifying an aggregation function for each graph. However, such a naive extension may lead to the loss of multimodal information, and potentially increase the storage and computational burden. In order to focus on correspondences between data modalities, a set of data transformations are necessary to increase the homogeneity of the data within and across subjects and modalities and remove portions of the data that can be assumed not to be mutually represented. 

Motivated by the fact that tensor analysis can effectively model a family of multi-relational data and potential factor characteristics \cite{pmlr-v119-lee20i, hung2012multilinear, kolda2009tensor}, we take advantage of tensor representation for an effective fusions of graphs with different modalities. Specifically, given a multimodal brain network dataset $\mathcal{D} = \{ (\{G_{1s}, \cdots, G_{ms}, \cdots, G_{Ms} \}, y_s) \}_{s = 1}^S$, we construct a single 4D tensor $\mathcal{X} \in \mathbb{R}^{N \times N \times M \times S}$ with weighted connectivity matrices $\mathbf{X}_{ms} = \mathcal{X}(:,:,m,s)$ for all $M$ modalities and $S$ subjects. To explore the uniformity of multimomal brain networks and also capture most data variation, we adopt the feature extraction method in \cite{rajwade2012image}, and introduce common feature projection matrices $\mathbf{U}_1$ and $\mathbf{U}_2$ by minimizing the following problem
\begin{equation}\label{eq:solve_U1_U2}
\begin{split}
 \min_{\mathbf{C}_{ms}, \mathbf{U}_1,\mathbf{U}_2} \quad & \sum_{m=1}^M \sum_{s = 1}^S \|\mathbf{X}_{ms} - \mathbf{U}_1 \mathbf{C}_{ms} \mathbf{U}_2^T\|_F^2 \\
 \textrm{s.t.} & \quad \mathbf{U}_1^T \mathbf{U}_1 = \mathbf{I} \quad \text{and} \quad \mathbf{U}_2^T \mathbf{U}_2 = \mathbf{I}
\end{split}
\end{equation}
where $\mathbf{U}_1$ characterizes node-level relationship, $\mathbf{U}_2$ is used for feature extraction and $\mathbf{C}_{ms}$ is the coefficient matrix of $\mathbf{X}_{ms}$ obtained via $\mathbf{C}_{ms} = \mathbf{U}_1^T\mathbf{X}_{ms} \mathbf{U}_2$. Projection matrices $\mathbf{U}_1$ and $ \mathbf{U}_2$ can be obtained via the higher-order SVD (HOSVD) of tensor data $\mathcal{X}$ since Eq. (\ref{eq:solve_U1_U2}) can be written as
\begin{equation}\label{eq:solve_U1_U2_tensor}
\begin{split}
 \min_{\mathcal{C}, \mathbf{U}_1,\mathbf{U}_2} \quad & \|\mathcal{X} - \mathcal{C} \times _1\mathbf{U}_1 \times _2\mathbf{U}_2\|_F^2 \\
 \textrm{s.t.} & \quad \mathbf{U}_1^T \mathbf{U}_1 = \mathbf{I} \quad \text{and} \quad \mathbf{U}_2^T \mathbf{U}_2 = \mathbf{I}
\end{split}
\end{equation}
We notice that $\mathbf{U}_1 = \mathbf{U}_2$ due to the symmetric property of $\mathbf{X}_{ms}$, thus we are interested in the effectiveness of introducing $\mathbf{U}_1$. Intuitively, the left projection matrix $\mathbf{U}_1$ in Eq. (\ref{eq:solve_U1_U2}) captures the global node-level relationship, and one benefit of obtaining $\mathbf{U}_1$ with Eq. (\ref{eq:solve_U1_U2}) is that it does not require any prior knowledge such as node and graph labels, thus all data can be utilized efficiently. Also, when new subjects are available, $\mathbf{U}_1$ can be updated in an online fashion \cite{han2018online}. 
\vspace{-1pt}
\subsubsection{Multiplex GCN Aggregator}
An essential component of the propagation rule in Eq.~(\ref{eq:GCN}) is to define graph convolution filter in the spatial or spectral domain based on an aggregator, e.g., normalized adjacency matrix $\widetilde{\mathbf{A}}$. Unfortunately for brain network analysis under the multimodal environments, the common ROIs used to calculate $\widetilde{\mathbf{A}}$ \cite{zhang2018multi} are not always available in real-world cases. To address this issue, we notice that the $i$-th row of $\mathbf{U}_1$ encodes the weights for the $i$-th node in the projection $\mathbf{U}_1^T \mathbf{X}_{ms}$. Hence, by selecting the important columns of $\mathbf{U}_1$ according to the singular values \cite{lu2008mpca}, we could use the truncated $\mathbf{U}_1$ and $K$-Nearest Neighbor (KNN) graph to define an undirected adjacency matrix to facilitate the cross-modality learning for multiplex GCNs. To be specific, we identify the set of nodes $N_i$ that are neighbors to the node $v_i$ using KNN, and connect $v_i$ and $v_j$ if $v_i \in N_j$ or if $v_j \in N_i$. More specifically, we define the adjacency matrix $\mathbf{A}$ \cite{zhang2018multi} as
\begin{equation}\label{Adjacency_matrix}
{a}_{ij} = \left\{
             \begin{array}{lr}
             \text{exp}(-\frac{\|\mathbf{u}_i -\mathbf{u}_j\|^2}{2\sigma^2}) \quad \text{if} \, v_i \in N_j \, \text{or}  \, v_j \in N_i, &  \\
             0 \quad \quad \quad  \quad \quad \quad \quad \, \, \, \text{otherwise}. &
             \end{array}
\right.
\end{equation}
where $\mathbf{u}_i$ is the $i$-th row of truncated $\mathbf{U}_1$, and $\sigma$ is the kernel width parameter. Then we can substitute $\widetilde{\mathbf{A}} = \mathbf{A} + \mathbf{I}$ with $\widetilde{d}_{ii} = \sum_j \widetilde{a}_{ij}$ in Eq.~(\ref{eq:GCN}). \\
\indent For simplicity, let us consider the first layer of Eq.~(\ref{eq:GCN}) with $\mathbf{H}^{(0)}_{ms} = \mathbf{C}_{ms}$ as input based on Eq.~(\ref{eq:solve_U1_U2}), and let $\hat{\mathbf{A}} = \widetilde{\mathbf{D}}^{-\frac{1}{2}} \widetilde{\mathbf{A}} \widetilde{\mathbf{D}}^{-\frac{1}{2}}$, then in our case, the propagation rule of each input feature matrix $\mathbf{C}_{ms} = \mathcal{C}(:,:,m,s)$ is
\begin{equation}\label{eq:prop_rule_Cij_projection}
\mathbf{H}^{(1)}_{ms} = \sigma(\hat{\mathbf{A}} \mathbf{C}_{ms} \mathbf{W}^{(0)}),  \, m \in [1, M], \, s\in [1,S]
\end{equation}
which could also be written as
\begin{equation}\label{eq:prop_rule_Xij_projection}
\mathbf{H}^{(1)}_{ms} = \sigma(\hat{\mathbf{A}} \mathbf{U}_1^T \mathbf{X}_{ms}\mathbf{U}_2\mathbf{W}^{(0)})
\end{equation}
where both $\mathbf{U}_2$ and the shared weight matrix $\mathbf{W}^{(0)}$ are used for feature extraction, and $\mathbf{W}^{(0)}$ is obtained in an end-to-end fashion, thus in practice, $\mathbf{U}_2$ and $\mathbf{W}^{(0)}$ can be combined to save some computing time. According to the symmetry of $\hat{\mathbf{A}}$, we can rewrite Eq.~(\ref{eq:prop_rule_Cij_projection}) as
\begin{equation}\label{eq:prop_rule_Xij_projection_simplified_tensor_projection}
\begin{split}
  \mathbf{H}^{(1)}_{ms} = &~\sigma(\hat{\mathbf{A}}^T \mathbf{C}_{ms} \mathbf{W}^{(0)})\\
  = &~\sigma(\mathbf{C}_{ms} \times_1 \hat{\mathbf{A}}^T \times_2 \mathbf{W}^{(0)^T})
\end{split}
\end{equation}
where $\times_i$ denotes the $i$-th mode product. Rearranging Eq.~(\ref{eq:prop_rule_Xij_projection_simplified_tensor_projection}) in the tensor form, the propagation rule for all graphs at the $l$-th layer is formulated as
\begin{equation}\label{eq:prop_rule_tensor_representation}
  \mathcal{H}^{(l+1)} = \sigma(\mathcal{H}^{(l)} \times_1 \hat{\mathbf{A}}^T \times_2 \mathbf{W}^{(l)^T})
\end{equation}
where $\mathcal{H}^{(0)} = \mathcal{C}$ and $\mathcal{H}(:,:,m,s) = \mathbf{H}_{ms}$.
\subsubsection{Modality Pooling}
For subject $s$ with $M$ modalities, the final output of Eq.~(\ref{eq:prop_rule_tensor_representation}) are $M$ feature matrices $\{\mathbf{H}_{1s}^{(L)}, \cdots, \mathbf{H}_{ms}^{(L)}, \cdots, \mathbf{H}_{Ms}^{(L)}\}$, where $\mathbf{H}_{ms}^{(L)} \in \mathbb{R}^{N \times D_{out}}$ corresponds to the $m$-th modality of subject $s$ in the $L$-th layer, and $D_{out}$ represents the output feature size. To integrate information of all $M$ modalities, we add an additional modality pooling layer by introducing a 1D trainable modality importance weight $\boldsymbol{\alpha} = \{\alpha_1, \cdots, \alpha_m, \cdots, \alpha_M\}$. The final feature embedding matrix $\mathbf{F}_s \in \mathbb{R}^{N \times D_{out}}$ for the $s$-th subject is calculated as a weighted combination of all modalities by
\vspace{-5pt}
\begin{equation}\label{eq:modality_combination}
  \mathbf{F}_s = \sum_{m = 1}^M \alpha_m \mathbf{H}_{ms}^{(L)}
\end{equation}
Notice that when all elements of $\boldsymbol{\alpha}$ are the same, then Eq. (\ref{eq:modality_combination}) boils down to the simple average pooling strategy used by the multi-view GCN (MVGCN) \cite{zhang2018multi}. Leveraging tensor representation and the propagation rule in Eq. (\ref{eq:prop_rule_tensor_representation}), we can write Eq. (\ref{eq:modality_combination}) as
\begin{equation}\label{eq:modality_combination_rewritten}
  \mathcal{F}_{out} = \sigma(\mathcal{H}^{(L-1)} \times_1 \hat{\mathbf{A}}^T \times_2 \mathbf{W}^{(L-1)^T}) \times_3 \boldsymbol{\alpha}^T
\end{equation}
where $\mathcal{F}_{out} \in \mathbb{R}^{N \times D_{out} \times S}$ is the output feature embeddings for all $S$ subjects with $\mathcal{F}_{out}(:,:,s) = \mathbf{F}_s$. Furthermore, if our model contains only one GCN layer with $L = 1$, then $\mathcal{F}_{out}$ can be obtained by
\begin{equation}\label{eq:modality_combination_one_layer}
\begin{split}
  \mathcal{F}_{out} & = \sigma(\mathcal{C} \times_1 \hat{\mathbf{A}}^T \times_2 \mathbf{W}^T) \times_3 \boldsymbol{\alpha}^T \\
  & = \sigma(\mathcal{X} \times_1 (\mathbf{U}_1\hat{\mathbf{A}})^T \times_2 (\mathbf{U}_2\mathbf{W})^T) \times_3 \boldsymbol{\alpha}^T
\end{split}
\end{equation}
Our model is efficient in that it avoids directly operating on long vectors, and contains only a few trainable parameters. Therefore, it can be effortlessly extended to large brain network datasets.
\subsubsection{FCN Layer}
Finally, a fully connected network (FCN) with \textit{softmax} is applied to output feature embeddings $\mathcal{F}$ for classification. It computes the probability distribution over the labels:
\begin{align}\label{eq:softmax}
    p(y_s=j|\mathbf{\mathbf{f}_s}) = {\text{exp} (\mathbf{w}^{\mathrm{T}}_j \mathbf{f}_s)}/{\left[\sum\nolimits_{k=1}^K \text{exp} (\mathbf{w}^{\mathrm{T}}_k \mathbf{f}_s)\right]},
\end{align}
where $\mathbf{w}_k$ is the weight vector of the $k$-th class, and $\mathbf{f}_s$ is the vectorized output feature embedding of subject $s$ obtained from $\mathcal{F}_{out}(:,:,s)$. Algorithm \ref{alg:MGNet} summarizes the main steps of the proposed {\ours} model.

\begin{algorithm}[t]
\small
\caption{Multiplex Graph Networks ({\ours})}
{\bf Input:} Dataset $\mathcal{X}$.\\
{\bf Step 1:} (Cross-Modality Bridging): 
Calculate the first mode projection matrix $\mathbf{U}_1$ of $\mathcal{X}$ according to Eq. (\ref{eq:solve_U1_U2}).\\
{\bf Step 2:} (Multiplex GCN Aggregator): 
Define the aggregated adjacency matrix $\widetilde{\mathbf{A}}$ by Eq.~(\ref{Adjacency_matrix}) using the truncated $\mathbf{U}_1$. Perform multi-layer GCNs according to the propagation rule in Eq. (\ref{eq:prop_rule_tensor_representation}).\\
{\bf Step 3:} (Modality Pooling and Prediction): 
Conduct modality pooling to extract feature embeddings for input subjects, which are then fed to FCN for classification. \\
{\bf Output:} class label $y$.\\
\label{alg:MGNet}
\vspace{-10pt}
\end{algorithm}

\section{Experiments} \label{sec:exp}
\subsection{Datasets and Preprocessing}
Table \ref{Table_dataset_description} lists the statistics of three real-world datasets used in our experiments, which are briefly introduced as below, and more details of data-preprocessing are provided in the supplementary material.\\
\textbf{Human Immunodeficiency Virus Infection (HIV)}: This dataset is collected from two modalities: functional magnetic resonance imaging (fMRI) and diffusion tensor imaging (DTI). The original dataset is heavily unbalanced, for the sake of exposition we randomly sampled 35 early HIV patients (positive) and 35 seronegative controls (negative). These two groups of subjects do not differ in demographic characteristics such as age, gender, racial composition and education level. \\
\textbf{Bipolar Disorder (BP)}: This dataset is collected across fMRI and DTI modalities, consisting of 52 bipolar I subjects in euthymia and 45 healthy controls with matched age and gender. The resting-state fMRI data was acquired on a 3T Siemens Trio scanner using a T2$^*$ echo planar imaging (EPI) gradient-echo pulse sequence with integrated parallel acquisition technique (IPAT) and DTI data were acquired on a Siemens 3T Trio scanner. \\
\textbf{Parkinson's Progression Markers Initiative (PPMI)}: This dataset is collected from three modalities: Probabilistic Index of Connectivity (PICo), Hough voting (Hough) and FSL. Similar to the HIV dataset, the original dataset is heavily unbalanced, so we randomly sample 149 patients with Parkinson's disease and 149 healthy controls.
\begin{table}[t]
  \centering
  \caption{Details of three datasets used in the experiments.}
  \scalebox{0.85}{
    \begin{tabular}{ccccc}
    \toprule
    Dataset & Class & \# Samples & Sample size & \ Modality \\
    \midrule
    \multirow{2}[4]{*}{HIV} & Health & 35    & \multirow{2}[4]{*}{90 x 90 x 2} & \multirow{2}[4]{*}{fMRI \& DTI} \\
\cmidrule{2-3}          & Patient & 35    &       &  \\
    \midrule
    \multirow{2}[4]{*}{BP} & Health & 45    & \multirow{2}[4]{*}{82 x 82 x 2} & \multirow{2}[4]{*}{fMRI \& DTI} \\
\cmidrule{2-3}          & Patient & 52    &       &  \\
    \midrule
    \multirow{2}[4]{*}{PPMI} & Health & 149   & \multirow{2}[4]{*}{84 x 84 x 3} & \multirow{2}[4]{*}{PICo \& Hough \& FSL} \\
\cmidrule{2-3}          & Patient & 149   &       &  \\
    \bottomrule
    \end{tabular}%
    }
  \label{Table_dataset_description}%
\end{table}%

\subsection{Baselines and Metrics}
To demonstrate the effectiveness of our {\ours} model, we compare it with the following up-to-date strong baselines for graph and neurological disorder analysis on the above multimodal HIV, BP and PPMI datasets.
\begin{itemize}
    \item \emph{\textbf{M2E}} \cite{liu2018multi}: It is a tensor-based method for multimodal feature fusion. We apply it to get the embeddings of all subjects and then perform classification with FCN.
    \item \emph{\textbf{MIC}} \cite{shao2015clustering}: It first uses the kernel matrices to form an initial tensor across all the multiple sources. After that, CP decomposition was employed to extract feature representation for each subject. We then perform classification using the same settings as above.
    \item \emph{\textbf{MPCA}} \cite{lu2008mpca}:
    We concatenate all data information into a 4D tensor and then apply MPCA to extract feature embeddings for each subject across modalities and individuals. We then perform classification using the same settings as above.
    \item \emph{\textbf{MK-SVM}}
   \cite{dyrba2015multimodal}: It is a multiple kernel learning method dependent of the SVM classifier, where the graph kernel is calculated as the weighted sum of single modality kernels.
    \item \emph{\textbf{3D-CNN}} \cite{gupta2013natural}: For each subject, we concatenate fMRI and DTI brain networks into a 3D tensor. Then we apply 3D-CNN for joint feature extraction and classification in an end-to-end manner.
   \item \emph{\textbf{GCN}}
   \cite{kipf2016semi}: We reshape the 4D tensor data $\mathcal{X} \in \mathbb{R}^{N \times N \times M \times S}$ into a 2D feature matrix $\mathbf{X} \in \mathbb{R}^{S \times MN^2}$, where each row corresponds to the vectorized representation of 3D multimodal data $\mathcal{X}(:,:,:, i)$, in this case, each vectorized graph can be viewed as a node, and the GCN model can be directly applied.
   \item \emph{\textbf{GAT}}
   \cite{velivckovic2017graph}:
   Similar to GCN, we also vectorize the input tensor data into a 2D feature matrix $\mathbf{X} \in \mathbb{R}^{S \times MN^2}$, and apply the graph attention mechanism.
   \item \emph{\textbf{DiffPool}}
    \cite{ying2018hierarchical}: It is a hierarchical GCN method equipped with differentiable pooling for graph classification. Since it can only handle single modal data, we apply it to each modality independently and report the best result.
   \item \emph{\textbf{MVGCN}} \cite{zhang2018multi}: It is a multi-view GCN method, which requires prior knowledge of common geometric coordinate information to define shared feature space. Typically, such information is not available in multimodal data, thus we consider to obtain the shared feature space with the average of all brain networks across modalities and subjects, and then feed it into the MVGCN architecture.
\end{itemize}
\vspace{-2pt}
In order to measure the performance of all compared methods, we use Accuracy and Area Under ROC Curve (AUC) as objective indicators of classification effectiveness, which are two widely used evaluation metrics for disease identification in medical fields. The larger the values, the better the classification performance.
\subsection{Implementation Details}
For all our experiments, we use binary cross-entropy loss and smooth $L_1$ loss with Adam optimizer \cite{kingma2014adam} to train the deep models. We emperically set the learning rate to $0.001$ and the epoch to $50$ iterations. We vary the dropout rate of the graph embedding layer from 0 to 0.5, and the number of MGNet layers from $\{1, 2, 3\}$. In our proposed model, there are three major parameters, namely the batch size, the number of the $K$ nearest neighbors when building the KNN graph, and the output feature size $D_{out}$ in GCNs. We apply the grid search to determine the optimal values of these three parameters. In particular, we empirically select $D_{out}$ from $\{ 20, 40, 60, 80, 100, 120\}$, and $K$ and batch size from $\{2, 4, 6, 8, 10,12\}$. In our experiments, $80\%$ of data are used for training, $10 \%$ for validation and the rest $10 \%$ for testing. We also carefully tune parameters of all compared methods according to the authors' suggestions and report their best possible results using the same data splits and the same 10-fold cross-validation scheme for a fair comparison. All experiments are performed on a 8-core machine with 16GB RAM. The deep learning backend is Tensorflow-gpu (2.2.0) with Python 3.6.
\begin{table*}[htbp]
\small
  \centering
  \caption{Comparison of different methods on HIV, BP and PPMI datasets in terms of average Accuracy and AUC scores.}
    \begin{tabular}{cccccccc}
    \toprule
    \multirow{2}[4]{*}{Model} & \multirow{2}[4]{*}{Method} & \multicolumn{2}{c}{HIV} & \multicolumn{2}{c}{BP} & \multicolumn{2}{c}{PPMI} \\
\cmidrule{3-8}          &       & Accuracy & AUC   & Accuracy & AUC   & Accuracy & AUC \\
    \midrule
    \multirow{4}[8]{*}{Shallow} & M2E   & 50.61$\pm$15.84 & 51.53$\pm$13.68 & 57.78$\pm$12.61 & 53.63$\pm$11.82 & 55.93$\pm$8.34 & 54.79$\pm$7.32 \\
\cmidrule{2-8}          & MIC   & 55.63$\pm$15.28 & 56.61$\pm$13.43 & 51.21$\pm$13.78 & 50.12$\pm$16.78 & 56.03$\pm$8.26 & 55.21$\pm$6.86 \\
\cmidrule{2-8}          & MPCA  & 67.24$\pm$11.56 & 66.92$\pm$12.46 & 56.92$\pm$13.33 & 56.86$\pm$13.69 & 59.66$\pm$5.64 & 59.59$\pm$5.63 \\
\cmidrule{2-8}          & MK-SVM & 65.71$\pm$12.08 & 68.89$\pm$9.61 & 60.12$\pm$10.83 & 56.78$\pm$12.86 & 58.24$\pm$7.88 & 57.36$\pm$7.62 \\
    \midrule
    \multirow{5}[10]{*}{Deep}  & 3D-CNN & 74.31$\pm$18.81 & 73.53$\pm$16.41 & 63.33$\pm$11.21 & 61.62$\pm$10.26 & 63.89$\pm$7.12 & 66.18$\pm$7.23 \\
\cmidrule{2-8}          & GAT   & 68.58$\pm$13.51 & 67.31$\pm$14.32 & 61.31$\pm$15.04 & 59.93$\pm$13.54 & 63.45$\pm$6.93 & 63.43$\pm$7.25 \\
\cmidrule{2-8}          & GCN   & 70.16$\pm$12.54 & 69.94$\pm$12.91 & 64.44$\pm$15.71 & 64.24$\pm$16.45 & 63.10$\pm$7.98 & 63.09$\pm$8.10 \\
\cmidrule{2-8}          & DiffPool & 71.42$\pm$14.78 & 71.08$\pm$15.12 & 62.22$\pm$12.83 & 62.54$\pm$13.41 & 61.12$\pm$6.79 & 61.83$\pm$6.82 \\
\cmidrule{2-8}          & MVGCN & 74.29$\pm$11.27 & 73.75$\pm$12.63 & 62.22$\pm$16.83 & 62.64$\pm$16.89 & 59.24$\pm$7.83 & 58.38$\pm$7.92 \\
    \midrule
    Shallow + Deep & MGNet & \textbf{81.39$\pm$13.41} & \textbf{82.08$\pm$14.81} & \textbf{67.78$\pm$12.28} & \textbf{66.31$\pm$10.24} & \textbf{66.62$\pm$7.87} & \textbf{66.96$\pm$7.98} \\
    \bottomrule
    \end{tabular}%
   \label{Table_all_exp_results}%
   \vspace{-6pt}
\end{table*}%

\subsection{Results and Discussions}
To evaluate the effectiveness of our model, we conduct a series of experiments on the task of multimodal brain network classification to answer the following three questions: \textbf{Q1}: The performance of {\ours} compared with other state-of-the-art methods. \textbf{Q2}: The effectiveness of multiple modality learning in {\ours}. \textbf{Q3}: The benefits of combining tensor representation and GCN in our case.
\subsubsection{Model Comparison}
Table \ref{Table_all_exp_results} lists the results of all methods on HIV, BP and PPMI datasets. Notice that the first two datasets used in our experiments consist of less than 100 subjects, which leads to high standard deviations in the results. From Table \ref{Table_all_exp_results}, we have the following observations. \\
\indent Overall, it can be seen that the proposed MGNet yields an average improvement of 7.08 $\%$, 3.34$\%$ and 2.73$\%$  in terms of accuracy on classification tasks compared to the strong baseline methods on HIV, BP and PPMI dataset, respectively. Specifically, MGNet achieves more than 5$\%$ improvements compared with other graph-based models such as MVGCN and DiffPool on all three datasets, which demonstrates the effectiveness of the proposed embedding and common space projection strategy when dealing with multi-modality brain networks without ROIs. The advantage of MGNet over shallow tensor-based methods such as M2E and MPCA demonstrates the importance of modeling graph structures. Furthermore, compared with the straightforward GCN and GAT implementations, the superiority of the proposed MGNet lies mainly in the joint modeling of node- and modality-level relationship, which also preserves the multimodal tensor structure during node embedding and feature extraction.\\
\indent Furthermore, it is noticed that the 3D-CNN model produces very competitive results on all three datsets because it benefits from the locality features of the input data. However it obtains the graph representation flatly with convolution and pooling layers, which may not explicitly taking multimodal relationship and graph structure information into consideration. Besides, instead of training a multi-layer CNN model, the proposed MGNet provides an alternative solution with graph and tensor representation. \\
\indent To sum up, by taking advantage of the tensor representation and multiplex GCNs, MGNet is able to encode both the multimodal characteristics and capture graph structures of the input brain networks.
\subsubsection{Hyperparameter Analysis}
\begin{figure*}[htbp]
\graphicspath{{Figs/}}
\centering
\subfigure[Modality weight $\boldsymbol{\alpha}$]{
\includegraphics[width=1.988in, height=1.338in]{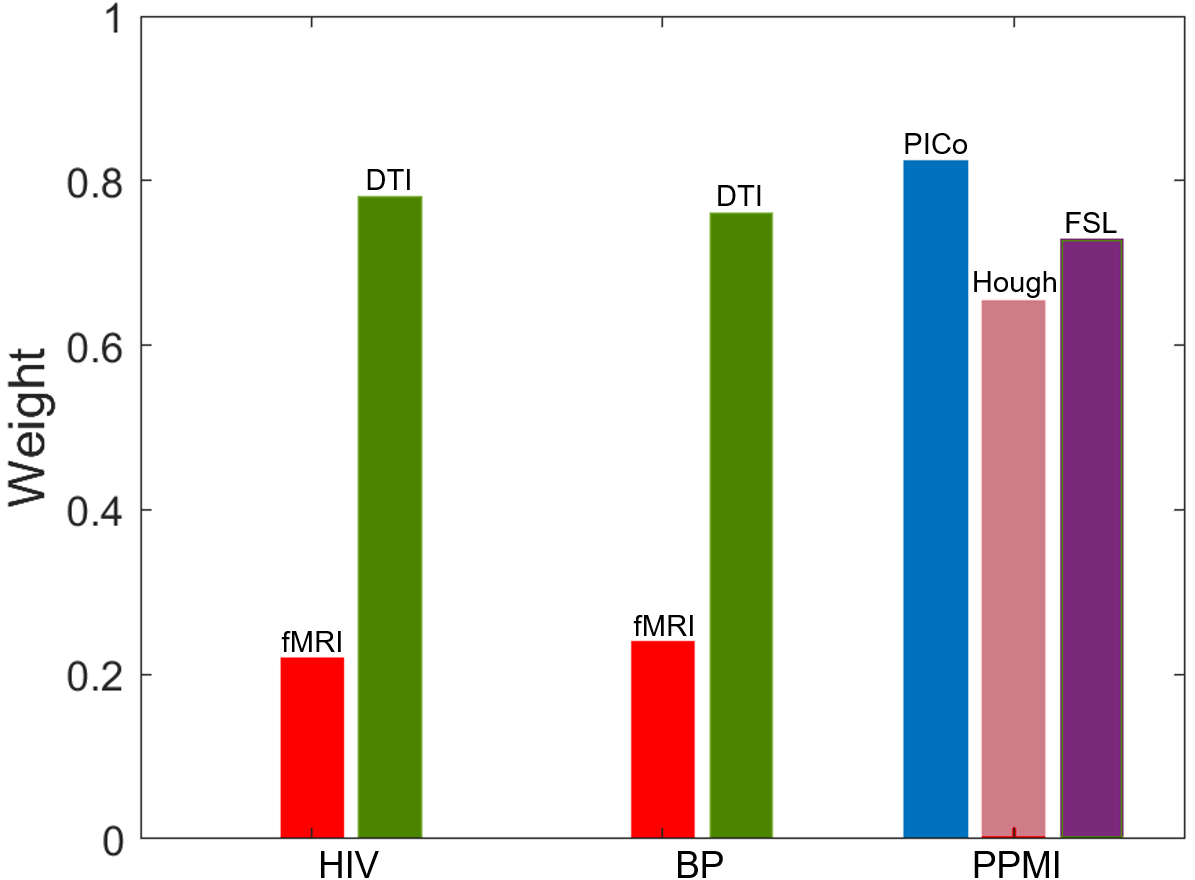}}
\subfigure[$\#$ Neighbours]{
\includegraphics[width=1.518in, height=1.318in]{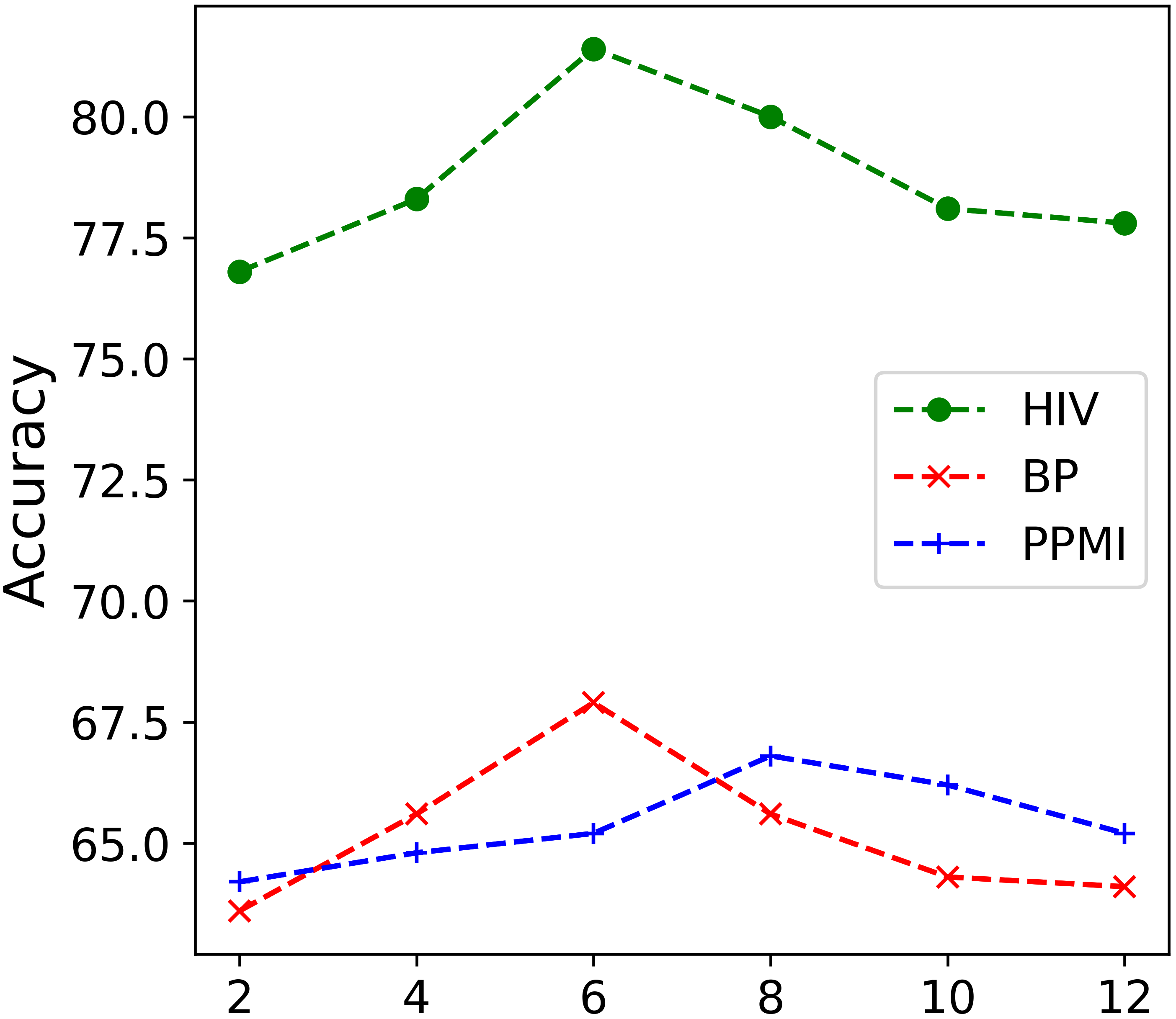}}
\subfigure[Feature size]{
\includegraphics[width=1.518in, height=1.318in]{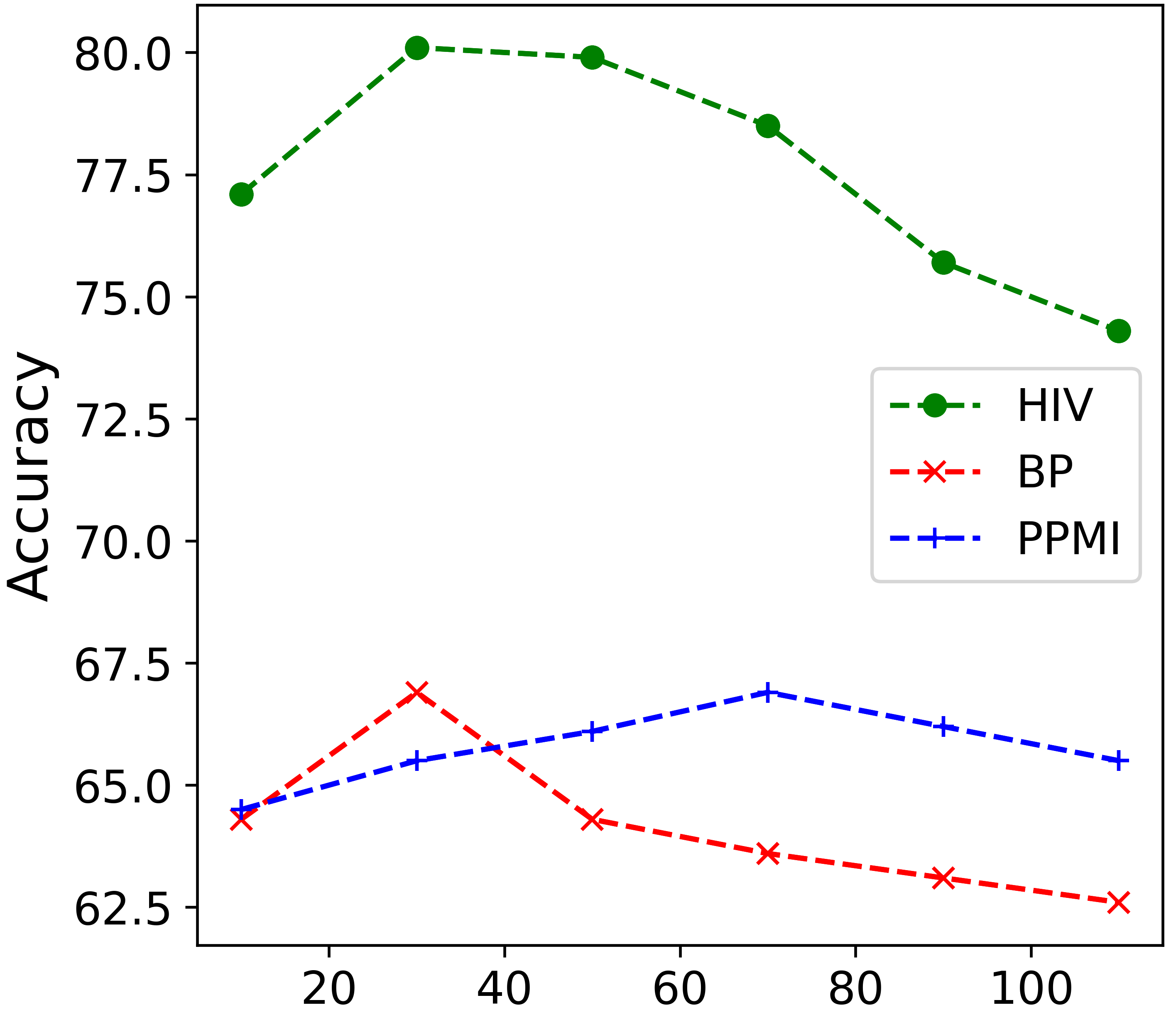}}
\subfigure[Batch size]{
\includegraphics[width=1.518in, height=1.31
in]{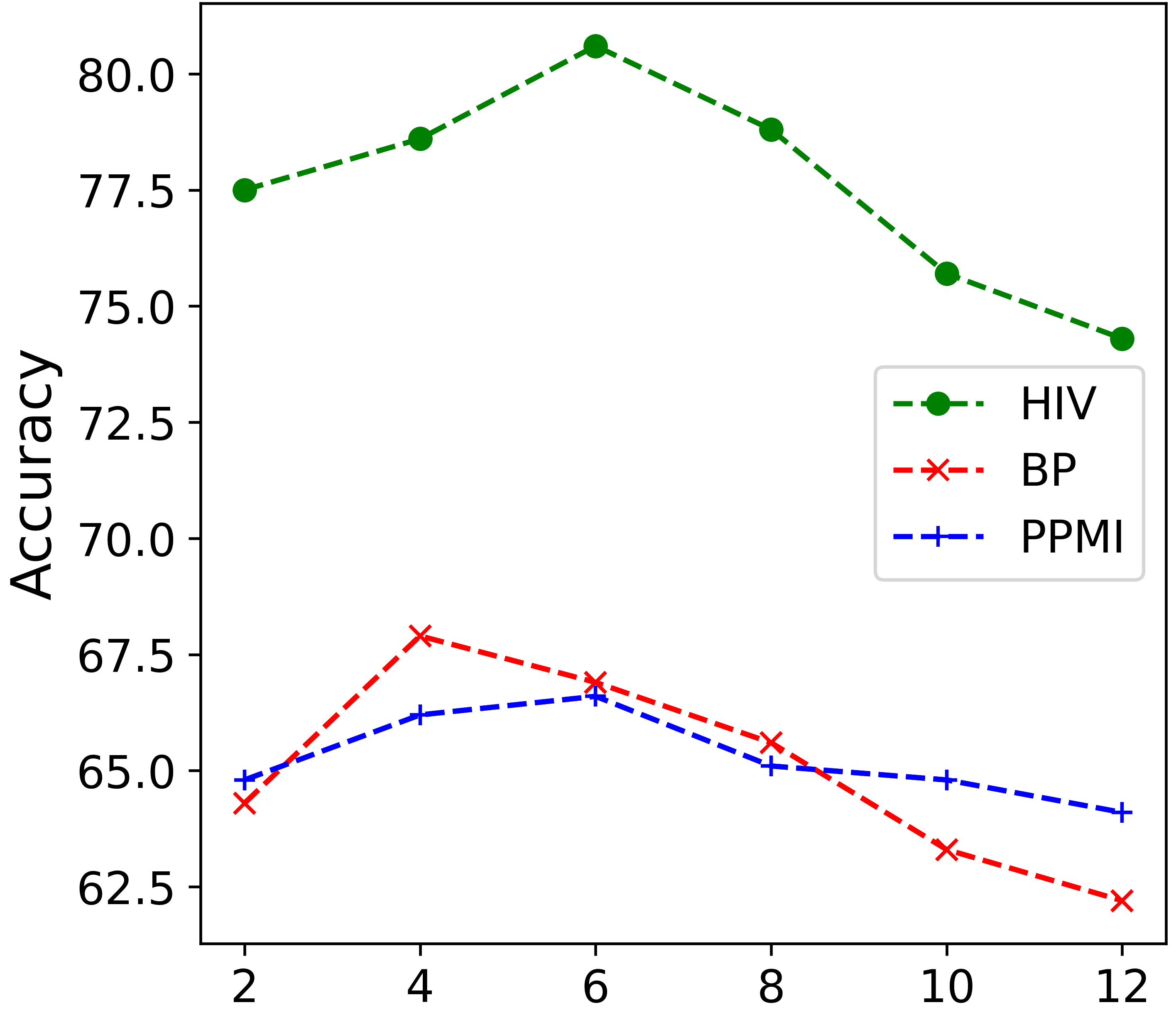}}
\caption{Modality weight $\boldsymbol{\alpha}$ and parameter sensitivity analysis of {\ours} on HIV, BP and PPMI datasets with varying numbers of neighbours, features and batch size.}
\label{Fig_parameter_tuning_and_weights}
\vspace{-6pt}
\end{figure*}

\noindent We investigate the effects of three important parameters in our {\ours} model, namely the number of neighbours $K$ when building the KNN graph, the dimensions of output features $D_{out}$ produced by GCN and the batch size used for training. According to Figs. \ref{Fig_parameter_tuning_and_weights}(b)-(d), we notice that the performance of {\ours} are related to all three parameters, which should be carefully tuned based on site-specific conditions.
\subsubsection{Effectiveness of Multimodal Learning}
In the proposed {\ours} model,  multi-modal information is integrated using the simple and effective modality-pooling strategy. To answer \textbf{Q2}, we investigate how the use of multiple modalities may improve the graph representation learning ability and thus the classification quality. Toward this end, we use HIV and BP datasets to compare {\ours} with their fine-tuned single-modality implementations. The classification results are listed in Table \ref{Table_ablation_view}.

\begin{table}[htbp]
\small
  \centering
  \caption{Results of MGNet with unimodal and multimodal data.}
  \scalebox{0.96}{
    \begin{tabular}{ccccc}
    \toprule
    Method & Dataset & Modality & Accuracy & AUC \\
    \midrule
    \multirow{6}[12]{*}{MGNet} & \multirow{3}[6]{*}{HIV} & fMRI   & 53.61$\pm$11.88 & 52.31$\pm$13.63 \\
\cmidrule{3-5}          &       & DTI  & 75.71$\pm$15.12 & 76.08$\pm$15.61 \\
\cmidrule{3-5}          &       & Both  & \textbf{81.39$\pm$13.41} & \textbf{82.08$\pm$14.81} \\
\cmidrule{2-5}          & \multirow{3}[6]{*}{BP} & fMRI   & 57.78$\pm$12.61 & 51.83$\pm$10.84 \\
\cmidrule{3-5}          &       & DTI  & 64.44$\pm$13.59 & 63.34$\pm$15.91 \\
\cmidrule{3-5}          &       & Both  & \textbf{67.78$\pm$12.28} & \textbf{66.31$\pm$10.24} \\
    \bottomrule
    \end{tabular}%
    }
  \label{Table_ablation_view}%
\end{table}%

From Table \ref{Table_ablation_view}, we can see that the multimodal learning strategy leads to significant improvements on both HIV and BP datasets. Owing to the usefulness of exploiting the complementary among multiple modalities or multiple types of features, the multimodality learning strategy is able to effectively improve the performance compared to the single-modality implementation. Another interesting observation is that our model with the DTI modality produces much better results compared to the fMRI modality, hence it is assigned a higher weight in the modality pooling operator $\boldsymbol{\alpha}$, as illustrated in Fig. \ref{Fig_parameter_tuning_and_weights} (a).
Furthermore, comparing the results of Table \ref{Table_all_exp_results} and Table \ref{Table_ablation_view}, we notice that CNN produces similar performance as the single-modality implementation of our MGNet model, which also implies the advantage of explicitly characterizing graph structures and multimodal correlation.
\subsubsection{Ablation Study}
In the formula of MGNet (Eq.~(\ref{eq:modality_combination_one_layer})), $\mathbf{U}_{1}$ captures the node-level information while $\boldsymbol{\alpha}$ characterizes the multimodal relationship, thus it is interesting to investigate the influence of different projection matrices of our MGNet model. In Table \ref{Table_mode_projection}, we compare MGNet with two different variations: MGNet without $\mathbf{U}_1$ (MGNet - $\mathbf{U}_1$) and MGNet with predefined average modality pooling $\boldsymbol{\alpha} = [0.5, 0.5]$ (MGNet + avg. Pooling). Comparing MGNet and MGNet - $\mathbf{U}_1$, it can be seen that the introduction of the node projection matrix $\mathbf{U}_1$ significantly boosts the performance of our model.

\begin{table}[htbp]
\small
  \centering
  \caption{Average accuracy of three different MGNet implementations on HIV, BP and PPMI datasets.}
  \scalebox{0.96}{
    \begin{tabular}{cccc}
    \toprule
    dataset & MGNet & MGNet+avg. Pooling  & MGNet - U1  \\
    \midrule
    HIV   & \textbf{81.39$\pm$13.41} & 80.81$\pm$12.04 & 75.71$\pm$14.33 \\
    \midrule
    BP    & \textbf{67.78$\pm$12.28} & 66.68$\pm$10.53 & 63.33$\pm$12.52 \\
    \midrule
    PPMI  & \textbf{66.62$\pm$7.87} & 65.17$\pm$8.14 & 62.78$\pm$7.05 \\
    \bottomrule
    \end{tabular}%
    }
  \label{Table_mode_projection}%
  \vspace{-6pt}
\end{table}%

To further investigate the role of $\mathbf{U}_1$, we visualize the difference between $\mathbf{X}_{ms}$ and $\mathbf{U}_1^T \mathbf{X}_{ms}$ for two subjects from the HIV brain network dataset in Fig. \ref{Fig_visual_projected_subjects}. An interesting observation is that although the two subjects belong to two different categories, the projected data $\mathbf{U}_1^T \mathbf{X}_{ms}$ exhibit similar patterns, where the signals of different intensity are distributed in certain rows and columns, thus, it is also reasonable to assume that they may share the same node-level self-similarity modeled by a uniform aggregation function in Eq.~(\ref{eq:modality_combination_rewritten}).

\begin{figure}[htbp]
\centering
\subfigure[Health]{
\includegraphics[width=1.57in]{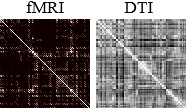}}
\subfigure[Patient]{
\includegraphics[width=1.57in]{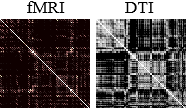}}
\subfigure[Health]{
\includegraphics[width=1.57in]{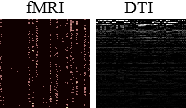}}
\subfigure[Patient]{
\includegraphics[width=1.57in]{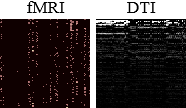}}
\caption{Comparison of original and projected data from the HIV dataset. The first row shows original data samples ($\mathbf{X}$). The second row illustrates corresponding projected data ($\mathbf{U}_1^T \mathbf{X})$.}
\label{Fig_visual_projected_subjects}
\vspace{-6pt}
\end{figure}

Interestingly, from Table \ref{Table_mode_projection}, we also notice that using a predefined modality pooling strategy $\boldsymbol{\alpha}$ could produce similar results. This can be explained by the fact that the DTI modality contains much more nonzero elements than the fMRI modality, as illustrated in Fig. \ref{Fig_visual_projected_subjects}, thus even if average view pooling is used, the DTI modality still dominates the training process.
\vspace{-3pt}
\subsubsection{Visualization}
The accuracy and quality of objective evaluations may be undermined by the very limited number of training and validation samples, also one important task of graph-based problems is to produce satisfactory representations for subsequent analysis purposes. Therefore, to qualitatively examine the effectiveness of MGNet, we use t-SNE \cite{maaten2008visualizing} to compare the graph embeddings learned by MVGCN and our MGNet on the HIV dataset, the results are illustrated in Fig.~\ref{Fig_visual_embeddings}.
\vspace{-3pt}

\begin{figure}[htbp]
\centering
\subfigure[Original]{
\includegraphics[width=1.45in, height=1.36in]{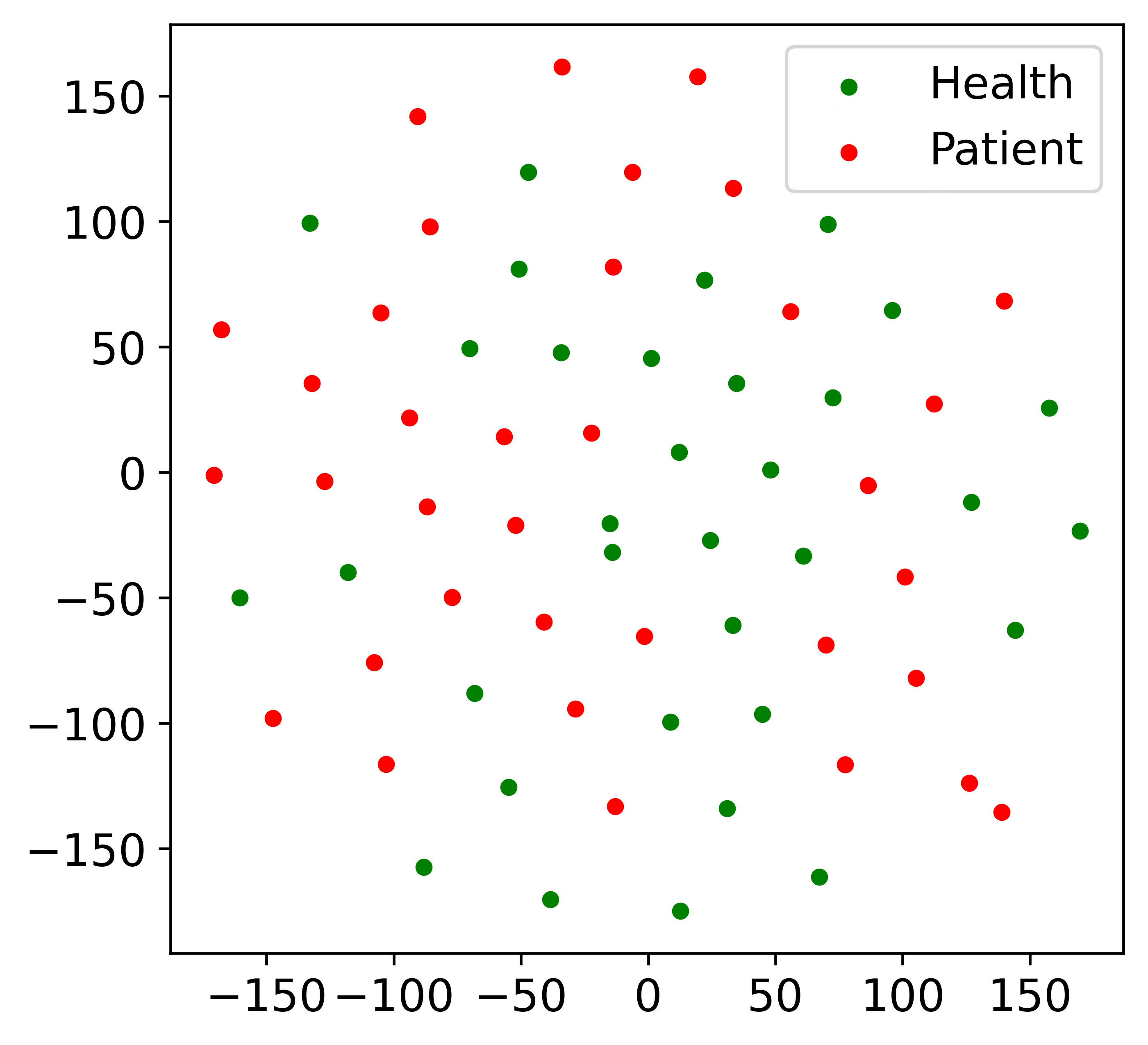}}
\subfigure[GCN]{
\includegraphics[width=1.45in, height=1.36in]{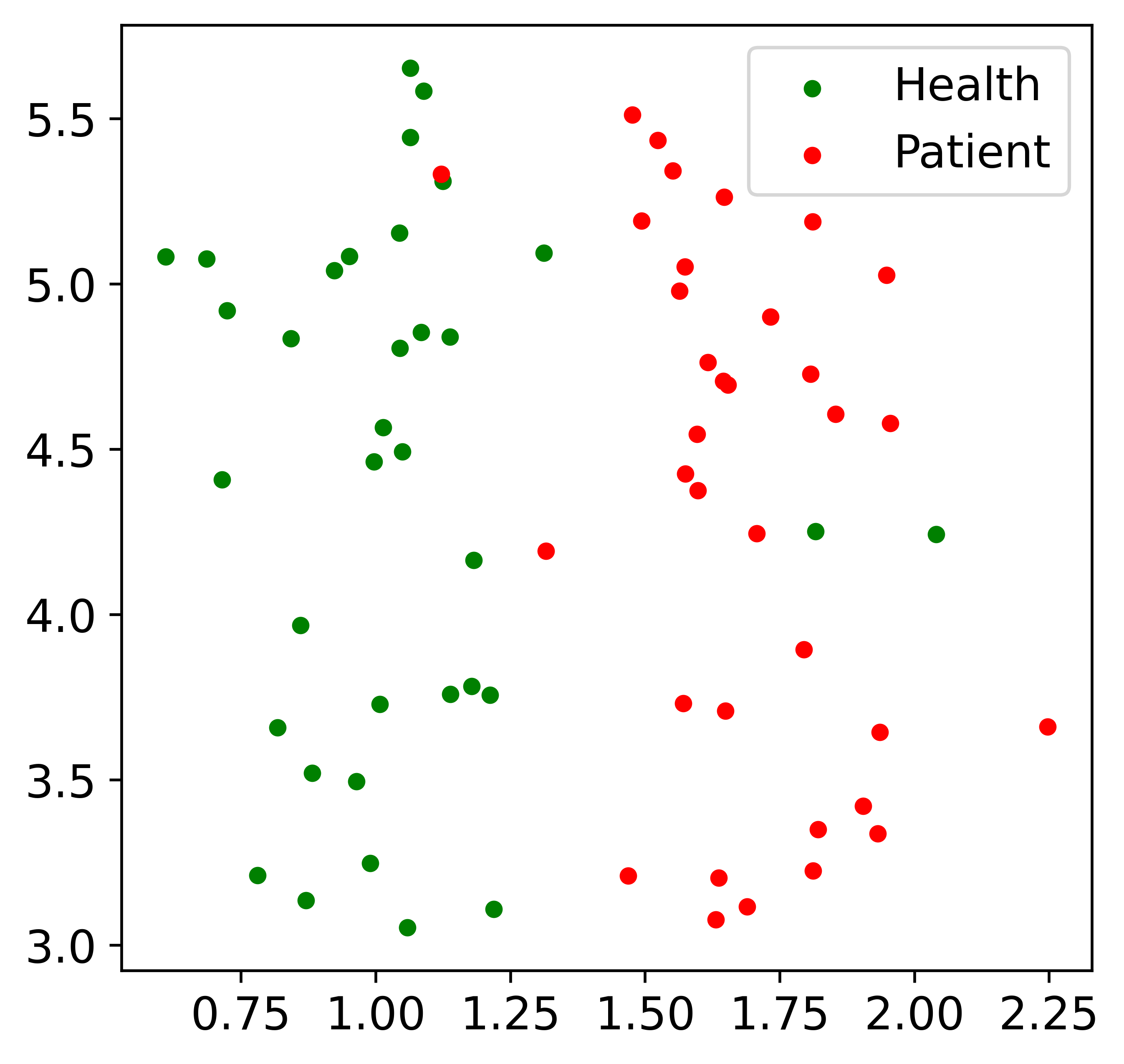}} \\
\subfigure[MVGCN]{
\includegraphics[width=1.45in, height = 1.36in]{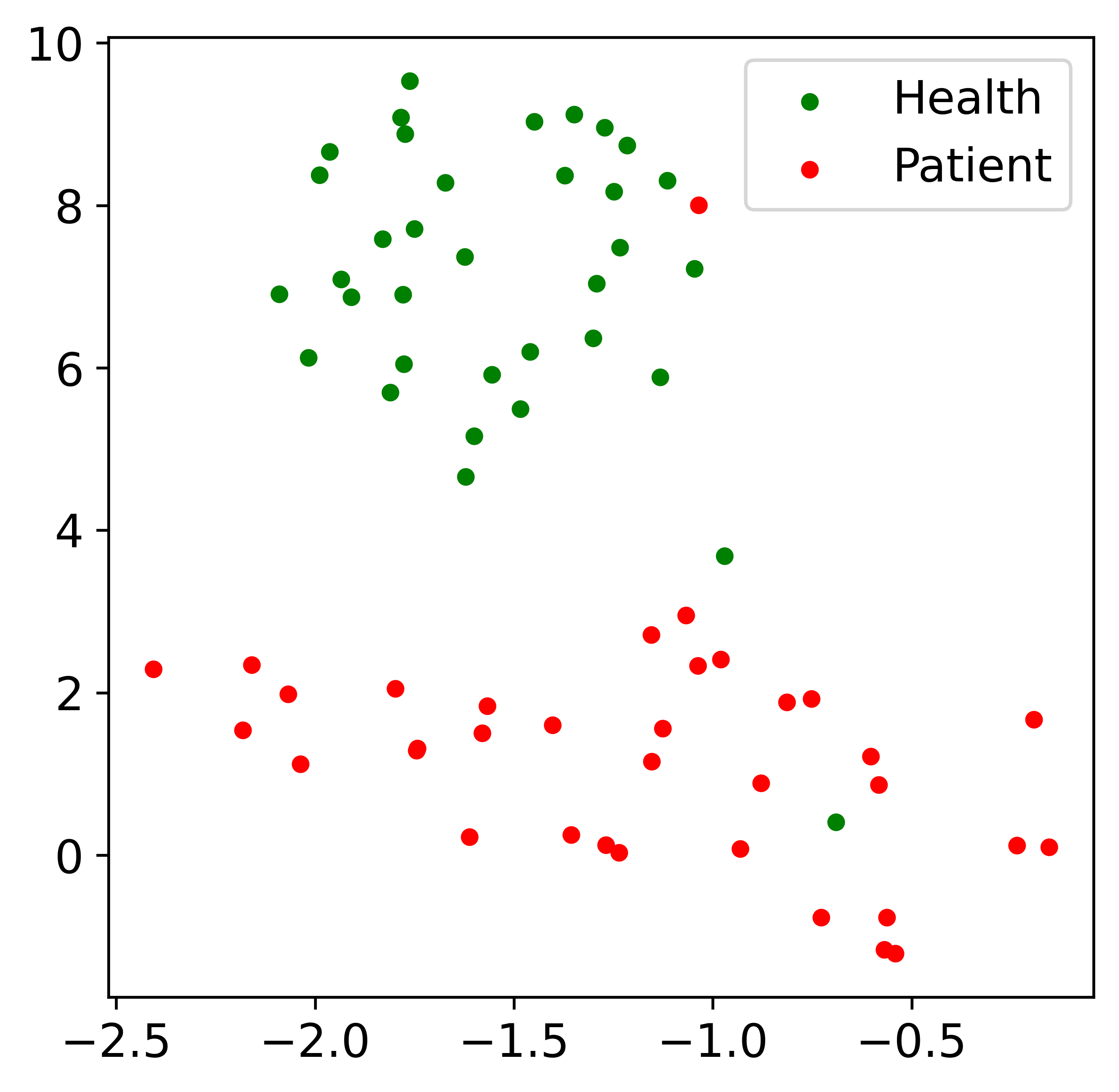}}
\subfigure[MGNet]{
\includegraphics[width=1.45in, height = 1.36in]{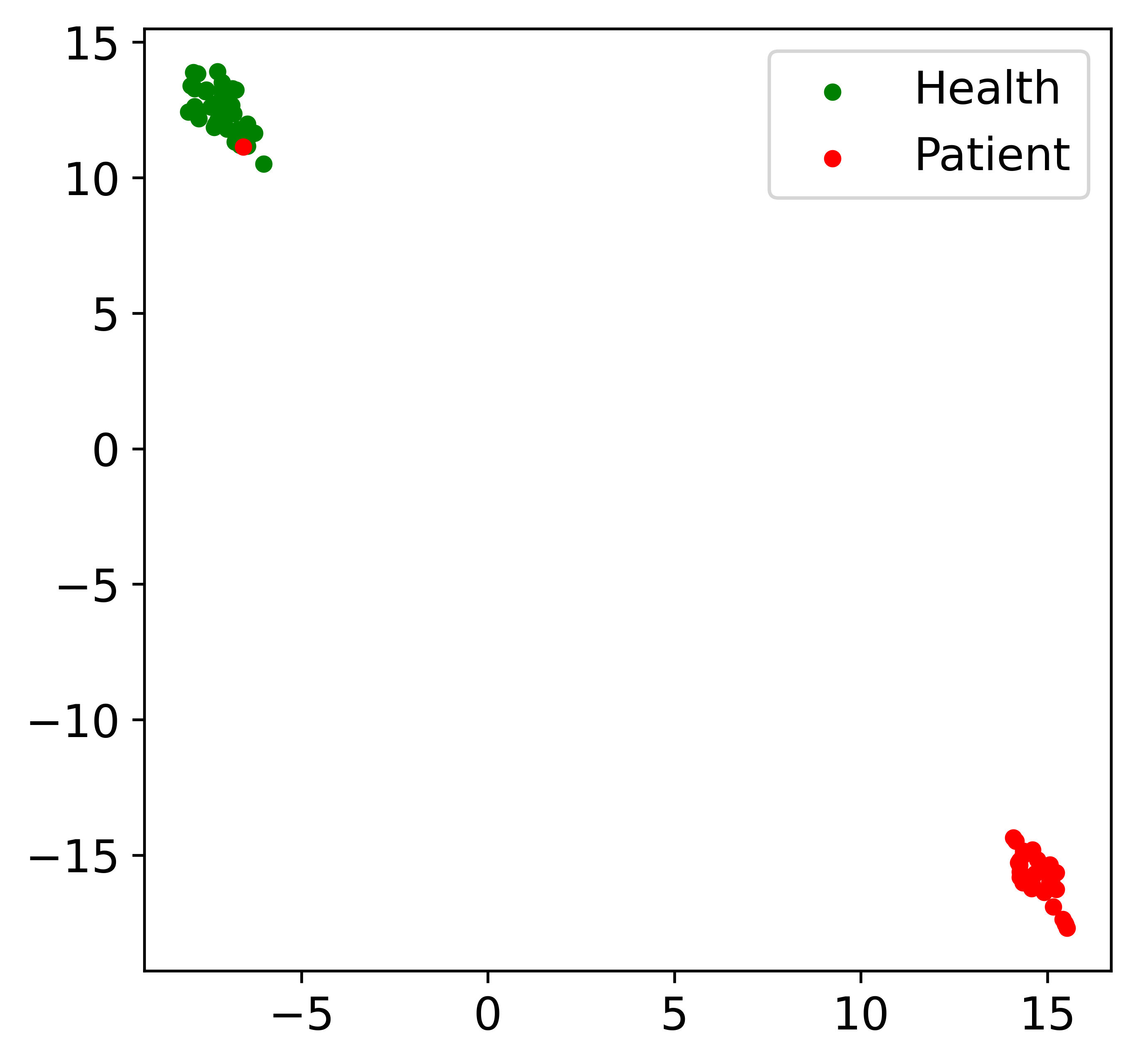}}
\caption{A tSNE visualization of the graph embeddings learned by GCN, MVGCN and MGNet on the HIV dataset.}
\label{Fig_visual_embeddings}
\vspace{-3pt}
\end{figure}
\noindent From Fig. \ref{Fig_visual_embeddings}, it can be seen that our MGNet model learns a higher quality of graph embeddings where the graphs are well-clustered according to their labels. \\
\indent To understand the performance of MGNet from clinical perspective, we also visualize the feature learning results on HIV and BP in Figs.~\ref{Fig_embed_node}. Left panel shows the node embedding features obtained by tensor decomposition. The coordinate system represents neuroanatomy and the color shows the activity intensity of the brain region. The right panel shows the graph embedding features which represent the factor strengths for both patients and healthy controls. From the left panel of Fig.~\ref{Fig_embed_node}, the embedded neuroanatomy is widely different from each other. From the right panel, the result shows that the controls have relatively positive correlation with node embedded feature, while the patient have relatively negative correlation. Moreover, those neuroimaging findings generally support clinical observations of functional impairments in attention, psychomotor speed, memory, and executive function. In particular, regions identified in our current study are consistent with those reported in structural and functional MRI studies of HIV associated neurocognitive disorder, including regions within the frontal and parietal lobes \cite{risacher2013neuroimaging}.

\begin{figure}[t]
\graphicspath{{Figs/}}
\centering
\subfigure[HIV]{
\label{Fig4}
\includegraphics[width=3.3in]{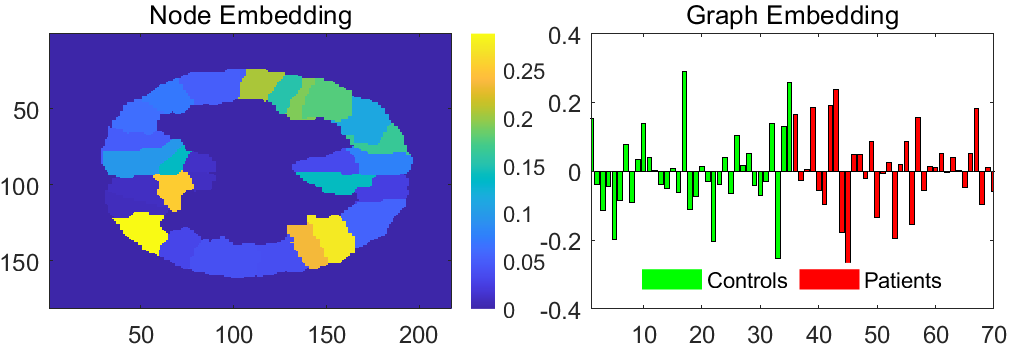}}
\subfigure[BP]{
\label{Fig4}
\includegraphics[width=3.3in]{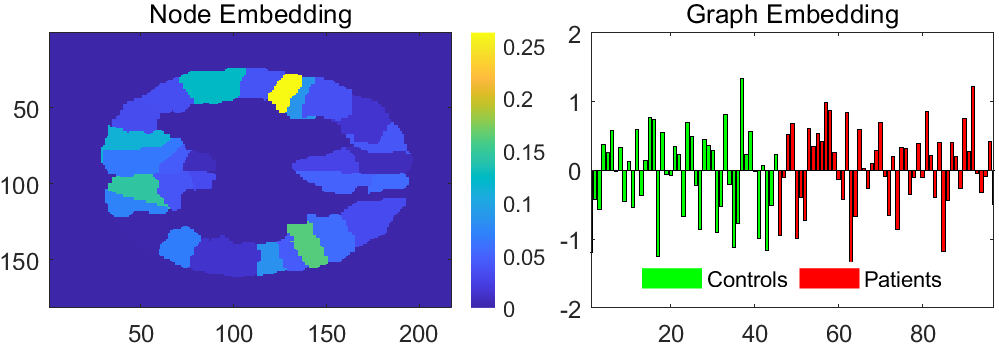}}
\caption{Embedded features of nodes and graphs of subjects (left and right panels respectively) from tensor decomposition and GCNs on HIV and BP datasets.}
\label{Fig_embed_node}
\vspace{-6pt}
\end{figure}


\section{Conclusions}\label{sec:conclude}
Multimodal brain network analysis is challenging in terms of both effectiveness and efficiency. Existing applications of GCN models explicitly requires a known graph structure (i.e., adjacency matrix), which is typically not available in multimodal cases. In this paper, we have presented a generic multiplex graph convolutional network ({\ours}) for multimodal brain network classification. It advances prior works by showing how tensor and GCN techniques can be combined together to effectively model multimodal graph-structured data for joint embedding and classification, without using any prior knowledge of the data. Compared with SOTA methods, {\ours} produces superior feature embedding and classification results on three realistic multimodal brain network datasets (i.e., HIV, Bipolar and PPMI). It is interesting to extend the proposed model to other tensor- and graph-based applications such as text analysis \cite{liu2020tensor}, dynamic graphs \cite{malik2021dynamic} and multi-relational learning \cite{ioannidis2020tensor}.

\bibliographystyle{ieee_fullname}
\bibliography{reference}

\end{document}